\documentclass{article}
\usepackage{spconf,amsmath,graphicx}
\usepackage{tabularx,booktabs}
\usepackage{subfigure}
\usepackage{algorithm,algorithmic}
\usepackage{url}

\title{SNOBERT: A Benchmark for clinical notes entity linking \\in the SNOMED CT clinical terminology}
\name{Mikhail Kulyabin$^{1}$,\quad Gleb Sokolov$^{2}$,\quad Aleksandr Galaida$^{2}$,\quad Andreas Maier$^{1}$,\quad Tomas Arias-Vergara$^{1}$}

\address{$^{1}$ Pattern Recognition Lab, University of Erlangen-Nuremberg, Germany \\ 
$^{2}$ Moscow Institute of Physics and Technology - State University (MIPT), Russia \\
}
\begin{document}
%\ninept
%
\maketitle
\begin{abstract}

The extraction and analysis of insights from medical data, primarily stored in free-text formats by healthcare workers, presents significant challenges due to its unstructured nature. Medical coding, a crucial process in healthcare, remains minimally automated due to the complexity of medical ontologies and restricted access to medical texts for training Natural Language Processing models. In this paper, we proposed a method, "SNOBERT," of linking text spans in clinical notes to specific concepts in the SNOMED CT using BERT-based models. The method consists of two stages: candidate selection and candidate matching. The models were trained on one of the largest publicly available dataset of labeled clinical notes. SNOBERT outperforms other classical methods based on deep learning, as confirmed by the results of a challenge in which it was applied.

%This paper addresses the Entity Linking problem, aims to link text spans in clinical notes to specific concepts in the SNOMED CT knowledge graph. 

\end{abstract}
\section{Introduction}
\label{sec:intro}

Most medical data is stored in free-text documents, usually filled in by healthcare workers. Analyzing this unstructured data can be challenging, as it can be difficult to extract meaningful insights. Medical coding remains an under-automated process despite being widely applicable in healthcare, medical insurance, and medical research, mainly due to the vast amount of codes in medical ontologies and the minimal access to medical texts for training natural language processing systems \cite{hristov2023clinical}. In recent years, the problem of Named Entity Recognition (NER) within medical texts has received increasing attention from the research community \cite{reyes2022clinical}. By applying standardized terminology, healthcare organizations can convert this free-text data into a structured format that computers can readily analyze, stimulating the development of new medicines, treatment pathways, and better patient outcomes. One of the most comprehensive and multilingual clinical healthcare terminologies in the world is Systematized Nomenclature of Medicine Clinical Terms (SNOMED CT) \cite{benson2012principles}, a systematically organized computer-processable collection of medical terms that provides codes, terms, synonyms, and definitions used in clinical documentation and reporting.

Annotating medical data according to SNOMED CT terminology is a time-consuming and labor-intensive process that often requires the annotator to have prior medical training. Automating such a process using Natural Language Processing (NLP) methods is called an Entity Linking (EL) problem. EL is the task of linking entities within a text to a suitable concept in a reference Knowledge Graph \cite{shen2014entity}. This work presents a "SNOBERT" method for linking text spans in clinical notes with specific topics in the SNOMED CT clinical terminology that distinguishes itself with a novel two-stage approach leveraging advanced NLP models and a refined preprocessing strategy. The method was applied to the "SNOMED CT Entity Linking Challenge" \cite{hardmansnomed}.

\section{Related Works}

With the recent advances of deep learning (DL) technologies, NLP applications have received an unprecedented boost in performance \cite{LAURIOLA2022443}. 
However, DL has rarely been used to solve the entity linking problem in SNOMED CT terminology since the massive size of the corpus needed to train on such a large set of classes \cite{gaudet2021use}. Nevertheless, few papers have been published in the field. Hristov et al. \cite{hristov2023clinical} proposed a method that integrates transformer-based models, such as BERT, pre-trained on biomedical data, with support vector classification using the transformer embeddings for fine-tuning and predicting SNOMED CT codes for medical texts. This hybrid approach leverages the strengths of both deep learning and classical machine learning techniques to achieve high accuracy in medical text coding, particularly in morphology and topography coding.

The KGE4SCT method \cite{CASTELLDIAZ2023104297} is a technique that utilizes Knowledge Graph Embeddings (KGEs) to automatically post-coordinate SNOMED CT clinical terms. It does this by using a vector space to capture the ontology's graph-like structure. The method uses vector similarity and analogies to derive post-coordinated expressions for clinical terms that are not explicitly present in SNOMED CT. This facilitates the encoding of clinical information that has been extracted from text. The effectiveness of this method has been validated on a subset of SNOMED CT and a set of manually post-coordinated concepts.

\label{sec:related}

\section{Data}
\label{sec:data}

In this work, we used the MIMIC-IV-Note dataset, which contains 331,794 de-identified hospital discharge summaries from 145,915 patients provided by the Beth Israel Deaconess Medical Center (BIDMC) and Massachusetts Institute of Technology (MIT) \cite{johnson2023mimic}. The challenge provided annotated data, comprising up to 300 annotated discharge summaries from the original MIMIC-IV-Note dataset. The full dataset consists of a public training subset (204 notes) and a private test subset (around 70 notes) comprising approximately 75,000 annotations across discharge summaries.

\begin{figure}[htb]
\begin{minipage}[b]{1.0\linewidth}
  \centering
  \centerline{\includegraphics[width=8cm]{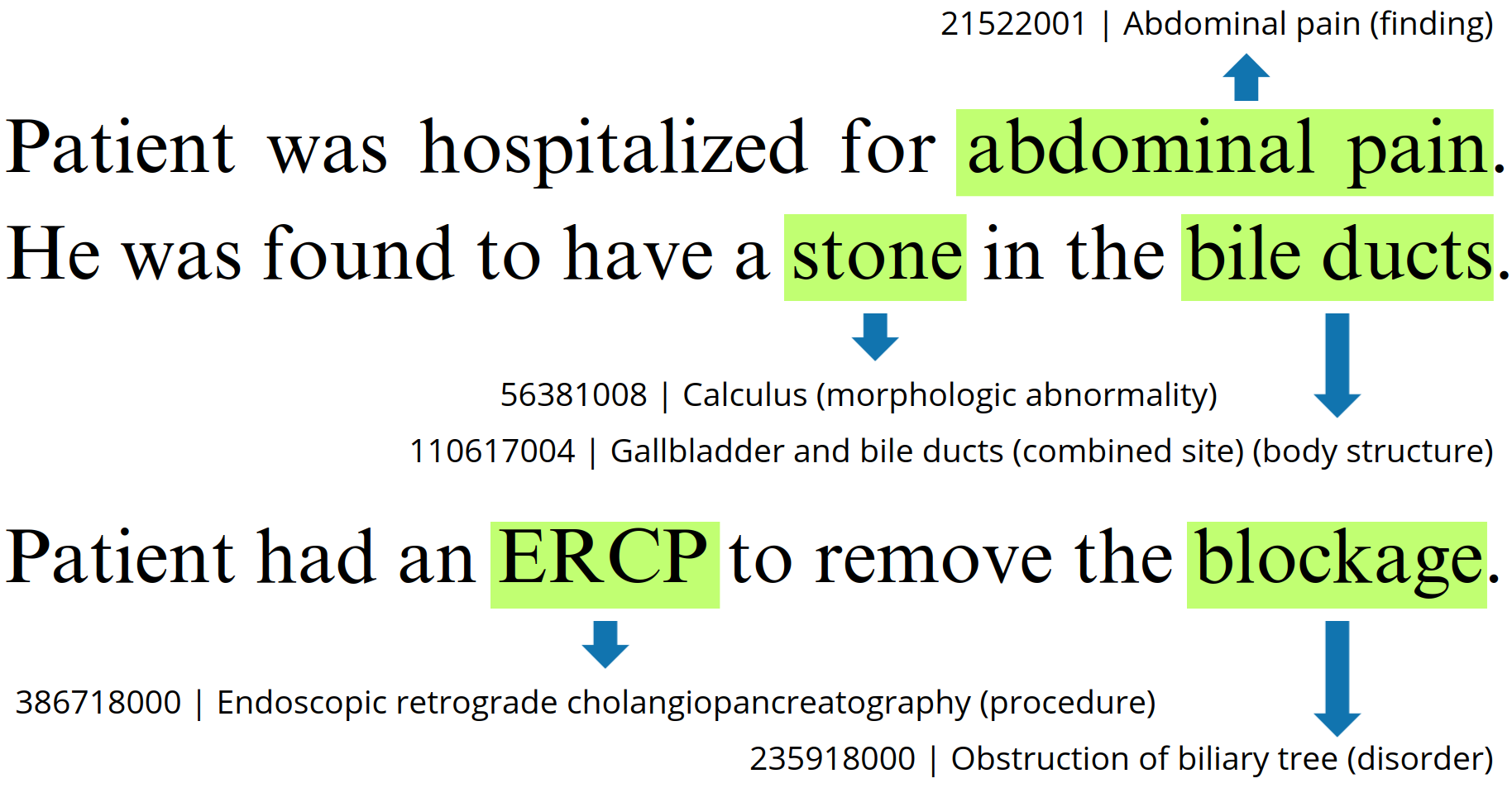}}
\end{minipage}
\caption{Example of synthetic discharge note annotated according to SMOMED CT terminology, e.g. "blockage" corresponds to "235918000" concept ID.}
\label{fig:snomed_ex}
\end{figure}

Each entry in the discharge dataset includes a note ID and the anonymized discharge text. Annotations indicate the concept ID, start and end points, and the corresponding note ID. Each note ID consists of a subject ID, note sequence position number, and note type (in this work, we use only discharges). Annotations consist of note IDs, starting and end points of concepts according to SNOMED CT clinical terminology, and their corresponding IDs. Fig.\ref{fig:snomed_ex} shows an example of an annotated part of a synthetic discharge note according to the SNOMED CT terminology.

\begin{figure}[htb]
\begin{minipage}[b]{1.0\linewidth}
  \centering
  \centerline{\includegraphics[width=6.5cm]{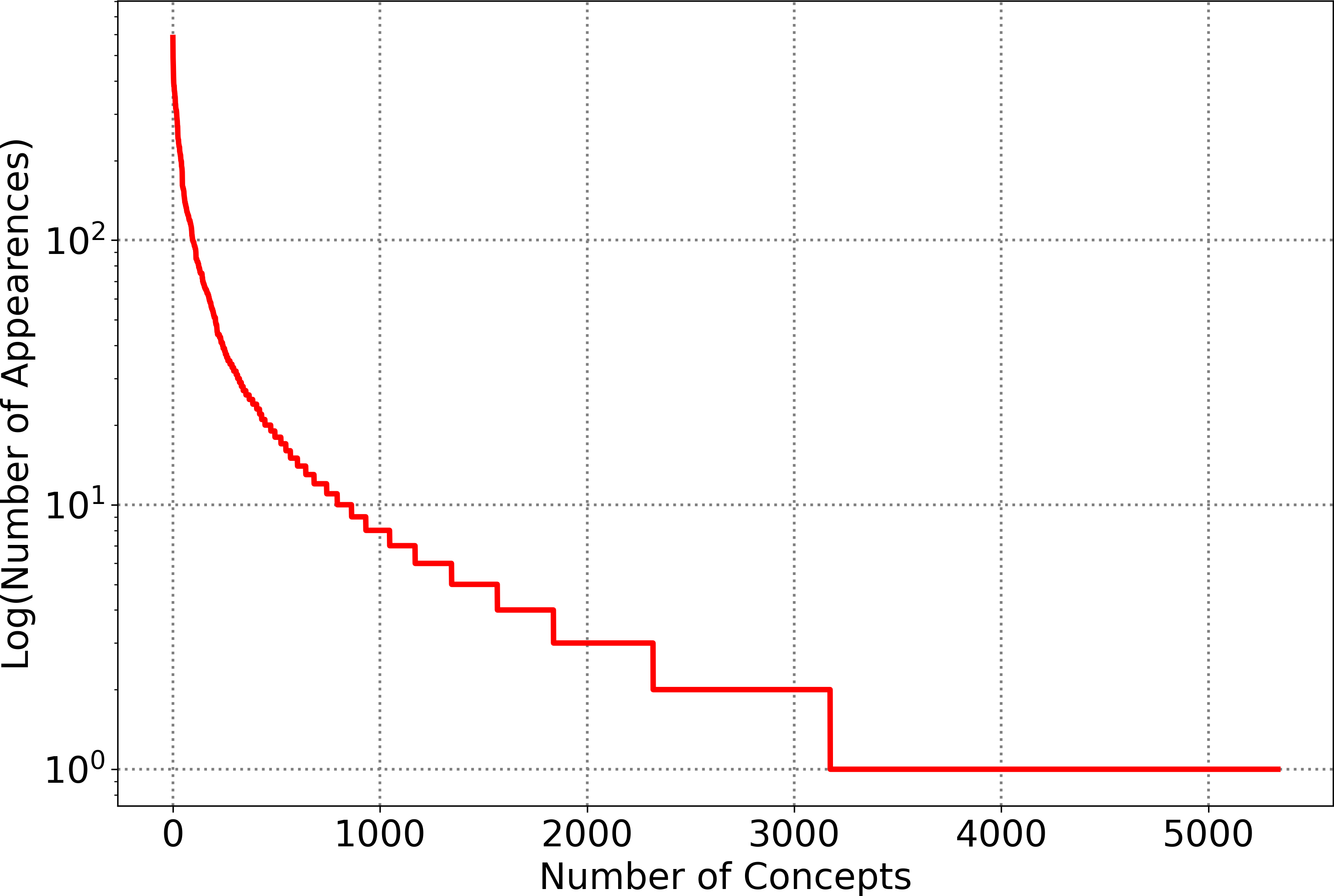}}
\end{minipage}
\caption{Distribution of the concepts in the annotated dataset: "long tail" distribution effect.}
\label{fig:tail}
\end{figure}

Medical notes frequently use abbreviations that can be context-dependent and may assume prior knowledge. In addition, the knowledge bases used in medical notes can contain hundreds of thousands of concepts, with many of these concepts occurring infrequently. As a result, there can be a "long tail" effect in the distribution of concepts, Fig.\ref{fig:tail}. Thus, 2162 concepts out of 5336 appear only once in the annotated data. This effect causes zero-shot learning (ZSL), a problem in deep learning in which, at test time, a learner observes samples from classes that were not observed during training and needs to predict the class to which they belong.

\section{Method}
\label{sec:method}

\begin{figure*}[h]
  \centering
  \includegraphics[width=17.5cm]{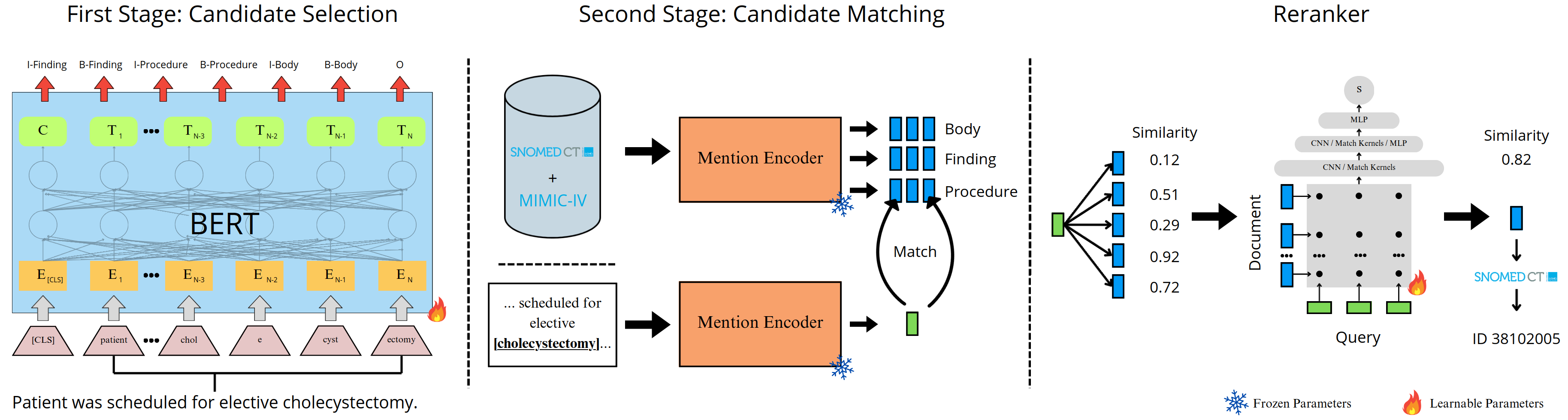}
\caption{SNOBERT scheme. The method consists of two stages. In the Candidate Selection stage (I), the BERT model is utilized to classify the text's tokens into seven classes. In the Candidate Matching stage (II), the Mention Encoder matches the extracted embeddings from the training and testing datasets within these classes. Reranker is used to rerank the top matches and to get the final similarity score.}
\label{fig:scheme}
\end{figure*}

This section presents the proposed method for the clinical notes EL. Fig.\ref{fig:scheme} illustrates the scheme of the method that consists of two stages: Candidate Selection and Candidate Matching. In the first stage, we solved the NER classification problem, and in the second stage, for each classified span from the first stage, we linked the corresponding concept ID in SNOMED terminology. In further sections, we describe each of the stages in detail.

\begin{figure}[htb]
\begin{minipage}[b]{1.0\linewidth}
  \centering
  \centerline{\includegraphics[width=5.4cm]{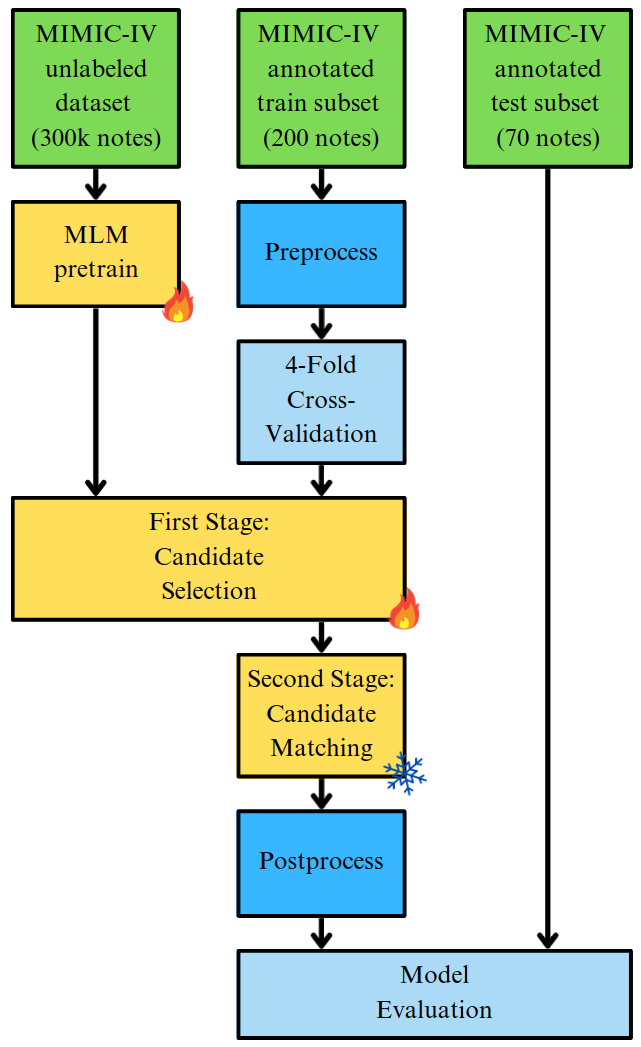}}
\end{minipage}
\caption{Training pipeline of the proposed approach. The method uses a two-stage solution: Candidate Selection and Candidate Matching. All the models were trained on the MIMIC-IV dataset. The model from the first stage was trained on the annotated training subset. An optional pretrain step was done on the full unlabeled dataset. Models were evaluated on the test annotated subset.}
\label{fig:overall}
\end{figure}

\subsection{Preprocessing}

We utilized the NER pipeline \cite{LabelStudio} to address certain annotation inaccuracies, such as those caused by shifts due to tags. Approximately 10 notes out of 204 underwent corrections, involving adjustments to around 150 annotation IDs. These corrections specifically targeted errors resulting from shifted annotations. Furthermore, most annotated notes are missing some labels in the paragraphs with the following headers: 'medications on admission:,' '\_\_\_ on admission:,' 'discharge medications:.' We excluded these parts from the training process. All HTML markup elements, such as the line break element ('br') or the new line ('n'), have also been removed from the notes.

While a robust baseline for EL, the dictionary method falls short when it comes to ZSL, highlighting the need for alternative solutions \cite{basaldella-etal-2020-cometa}. We generate a static dictionary of the most common concepts from training data and match them with test data in the post-processing step using a string-matching search. Levenshtein ratio or Stolois distance could be utilized as a matching metric. However, we applied "one-to-one" matching, linking only complete coincidences.

\subsection{First Stage: Candidate Selection}
NER is the task of identifying rigid designators' mentions from text belonging to predefined semantic types such as person, location, organization, etc. \cite{9039685}. NER involves processing raw text through stages, including Sentence Segmentation, where text is divided into sentences, and Word Tokenization, which breaks text into individual words. Subsequent stages include Part of Speech Tagging, assigning grammatical tags based on word roles and context, and Entity Detection, which identifies and categorizes key elements in the text, highlighting the core function of NER in extracting meaningful information from unstructured data.

We proposed to separate the concepts according to the first-level hierarchy (the first entry in the path to the concept) according to SNOMED terminology. Thus, we can emphasize the "Finding," "Procedure," "Body structure," and "None" classes. During Tokenization, the words got flags in "B-I-O" tagging format: "B" (Beginning) means the first token of the first word within an annotation; "I" (Inside) - is the first token of a subsequent word within an annotation; "O" (Outside) - is as a stand-alone token or single word token. Consequently, we have seven classes: "I-Finding," "B-Finding," "I-Procedure," "B-Procedure," "I-Body," "B-Body," and "O".

\subsection{Second Stage: Candidate Matching}
To link classified terms from the first stage, we match their embeddings with embeddings of terms from SNOMED terminology by cosine similarity. For this purpose, the whole database extracted concepts from the "Body structure," "Findings," and "Procedure" paths, which is about 200k unique IDs with the Mention Encoder. In this work, we applied the cambridgeltl/SapBERT-from-PubMedBERT-fulltext-mean-token model trained with UMLS 2020AA \cite{liu2021self}, using microsoft/BiomedNLP-PubMedBERT-base-uncased-abstract-fulltext as the base model \cite{pubmedbert}.

\subsection{Postprocessing}
A reranker is used to improve the performance of the model's initial predictions. The model generates a list of possible predictions, which might not be ranked optimally in order of correctness. A re-ranker evaluates these predictions and adjusts their ranking based on more refined or specific criteria. Using only the top one or top five predictions may cause the correctly predicted vector to be missed. In our method, as a re-ranker, we applied the MedCPT model \cite{jin2023medcpt} trained on 18M semantic query-article pairs from PubMed.

\subsection{Metrics}
We evaluated the proposed method on both stages. First stage was evaluated with Macro-F1 score across all labels:
\begin{equation} \label{eq:f1score}
Macro-F1 = \frac{1}{N} \sum^{N}_{i}{ \frac{2 \times Precision_{i} \times Recall_{i}}{Precision_{i} + Recall_{i}}},
\end{equation}
where $N$ is a number of classes.
%\begin{equation} %\label{eq:pre}
%Precision_{i} = \frac{TP_{i}}{TP_{i} + FP_{i}},
%\end{equation}
%\begin{equation} \label{eq:rec}
%Recall_{i} = \frac{TP_{i}}{TP_{i} + FN_{i}},
%\end{equation}

%$TP = True\ Positive$, $TN = True\ Negative$, $FP = False\ Positive$, $FN = False\ Negative$, and $N$ is a number of classes.

The second stage was evaluated with cosine similarity:

\begin{equation} \label{eq:cos}
cos(\theta) = \frac{ \sum^{n}_{i}{A_{i} B_{i}} }{ \sqrt{\sum^{n}_{i}{A^{2}_{i}}} \sqrt{\sum^{n}_{i}{B^{2}_{i}}}},
\end{equation}
where A and B are comparing vectors.

The final results were evaluated using a class macro-averaged character intersection-over-union (mIoU). IoU is a popular metric for measuring localization accuracy and computing localization errors. It calculates the amount of overlap between a prediction and a ground truth:

\begin{equation} \label{eq:iou}
IoU_{class} = \frac{P^{char}_{class} \cap G^{char}_{class}}{P^{char}_{class} \cup G^{char}_{class}},
\end{equation}

\begin{equation} \label{eq:iou}
macro \: IoU = \frac{\sum_{classes \in P \cup G} IoU_{class}}{N_{classes \in P \cup G}},
\end{equation}
where \(P_{class}^{char}\) is the set of characters in all predicted spans for a given class category, \(G_{class}^{char}\) is the set of characters in all ground truth spans for a given class category, and \(classes \in P \cup G\) are the set of categories present in either the ground truth or the predicted spans.

\subsection{Training}
For the training, we employed the 4-Fold Cross-Validation. Each fold consists of 51 discharge notes. For the first stage, we used a domain-specific pretrained language model for Biomedical Natural Language Processing \cite{https://doi.org/10.48550/arxiv.2112.07869}. We utilized the base version, microsoft/BiomedNLP-BiomedBERT-base-uncased-abstract-fulltext, which was pretrained on abstracts from PubMed and full-text articles from PubMedCentral and is available in the HuggingFace repository \cite{pubmedbert}. We ran four experiments for each setup so that three folds were used each time for training and one for validation. We trained each split on 100 epochs using EarlyStoppingCriteria. Therefore, 75 epochs on average were needed. We used ADAM optimizer with $3e^{-5}$, batch size of 8, and class weighting. Training takes 30 minutes on 4 GPUs (NVIDIA A100-SXM4-40GB). A more detailed description of the training configuration is shown in our training repository \cite{snobert}.

We found a slight improvement of 0.0005 in IoU when applying the Masked Language Model (MLM) pretraining technique. To accomplish this, we used the larger microsoft/BiomedNLP-BiomedBERT-large-uncased-abstract weights as an initial model instead. This optional pretraining step takes 24 hours on 4 GPUs.

\section{Results}

Table \ref{results} shows the averaged cross-validation evaluation results for the first (I) stage and final evaluation. MLM pretraining slightly improves the F1 score in the first stage, and therefore, mIoU in the second with the best score of 0.4302 training on 4 GPUs. MultiGPU outstands single due to the batch size and its synchronization in the Bert model.

\begin{table}[h]
\caption{Evaluation metrics of the proposed method}
\label{results}
\begin{tabular}{lclll}
\toprule
Model (I)  & GPUs & F1 (I) & mIoU & Epoch \\ 
\midrule
BiomedBERT large & 1 & 0.7429 & 0.4231 & 75 \\
BiomedBERT base & 1 & 0.7487 & 0.4199 & 76 \\
\textbf{BiomedBERT large} & 4 & \textbf{0.7514} & \textbf{0.4302} & 74 \\
BiomedBERT base & 4 & 0.7499 & 0.4257 & 72 \\
\bottomrule
\end{tabular}
\end{table}

Table \ref{results_2} shows the evaluation results of the second stage before reranking for each class. The results for the top five best candidates are significantly higher than those of the top one, which shows the need to use a reranker in the postprocess, as the best candidate after the second stage is not always at the top of the similarity score. 

\begin{table}[h]
\centering
\caption{Evaluation metrics of the second stage.}
\label{results_2}
\begin{tabular}{lcc}
\toprule
Category  & Similarity@1 & Similarity@5 \\ 
\midrule
Body structure & 0.715 & 0.811 \\
Findings & 0.614 & 0.703 \\
Procedure & 0.478 & 0.694 \\
\bottomrule
\end{tabular}
\end{table}

Table \ref{results_3} shows the final test results of the three best methods and the baseline of the competition. The dictionary-based method achieved the highest score. However, this is a time-consuming method that requires manual effort.

\begin{table}[h]
\centering
\caption{Results on the test dataset.}
\label{results_3}
\begin{tabular}{lcc}
\toprule
Author  & Method & mIoU \\ 
\midrule
Bilu et al. & Dictionary-based $^*$ & 0.4202 \\
Ours & SNOBERT & 0.4194 \\
Popescu et al. & Faiss + Mistral & 0.3777 \\
Baseline & deberta-v3-large & 0.1794 \\
\bottomrule
\end{tabular}

\smallskip

\parbox[t]{\textwidth}{\footnotesize
  \textit{$^*$}
Time-consuming semi-manual method.}    

\end{table}

% Results on the private test were obtained for pipelines with and without static dictionary. IoU for the method without the dictionary was 0.4159, with the dictionary 0.4194. The final metrics are slightly lower than those obtained with cross-validation, which can be explained by the large dataset size and heterogeneity of the data, but they are still close to claiming that the chosen approach with cross-validation was adequate.

\label{sec:results}

\section{Conclusion}
\label{sec:discussion}

We proposed a "SNOBERT" method of EL of text spans in clinical notes with specific topics in the SNOMED CT clinical terminology, tested in practice in the "SNOMED CT Entity Linking Challenge." Our method uses two stages; however, an end-to-end approach could improve the linking score \cite{ayoola-etal-2022-refined} but would need more training data from the other side. The limited annotation problem could be solved using synthetic data generated with Large Language Models \cite{kweon2023publicly}.

\bibliographystyle{IEEEbib}
\bibliography{refs}

\begin{thebibliography}{10}

\bibitem{hristov2023clinical}
Anton Hristov et~al.,
\newblock ``Clinical text classification to snomed ct codes using transformers
  trained on linked open medical ontologies,''
\newblock in {\em Proceedings of the 14th International Conference on Recent
  Advances in Natural Language Processing}, 2023, pp. 519--526.

\bibitem{reyes2022clinical}
Javier Reyes-Aguill{\'o}n et~al.,
\newblock ``Clinical named entity recognition and linking using bert in
  combination with spanish medical embeddings.,''
\newblock in {\em CLEF (Working Notes)}, 2022, pp. 341--349.

\bibitem{benson2012principles}
Tim Benson,
\newblock {\em Principles of health interoperability HL7 and SNOMED},
\newblock Springer Science \& Business Media, 2012.

\bibitem{shen2014entity}
Wei Shen, Jianyong Wang, and Jiawei Han,
\newblock ``Entity linking with a knowledge base: Issues, techniques, and
  solutions,''
\newblock {\em IEEE Transactions on Knowledge and Data Engineering}, vol. 27,
  no. 2, pp. 443--460, 2014.

\bibitem{hardmansnomed}
Will Hardman and others.,
\newblock ``Snomed ct entity linking challenge,''
\newblock 2024.

\bibitem{LAURIOLA2022443}
Ivano Lauriola, Alberto Lavelli, and Fabio Aiolli,
\newblock ``An introduction to deep learning in natural language processing:
  Models, techniques, and tools,''
\newblock {\em Neurocomputing}, vol. 470, pp. 443--456, 2022.

\bibitem{gaudet2021use}
Christophe Gaudet-Blavignac, Vasiliki Foufi, Mina Bjelogrlic, and Christian
  Lovis,
\newblock ``Use of the systematized nomenclature of medicine clinical terms
  (snomed ct) for processing free text in health care: systematic scoping
  review,''
\newblock {\em Journal of medical Internet research}, vol. 23, no. 1, pp.
  e24594, 2021.

\bibitem{CASTELLDIAZ2023104297}
Javier Castell-Díaz, Jose~Antonio Miñarro-Giménez, and Catalina
  Martínez-Costa,
\newblock ``Supporting snomed ct postcoordination with knowledge graph
  embeddings,''
\newblock {\em Journal of Biomedical Informatics}, vol. 139, pp. 104297, 2023.

\bibitem{johnson2023mimic}
Alistair~EW Johnson, Lucas Bulgarelli, Lu~Shen, Alvin Gayles, Ayad Shammout,
  Steven Horng, Tom~J Pollard, Sicheng Hao, Benjamin Moody, Brian Gow, et~al.,
\newblock ``Mimic-iv, a freely accessible electronic health record dataset,''
\newblock {\em Scientific data}, vol. 10, no. 1, pp. 1, 2023.

\bibitem{LabelStudio}
Maxim Tkachenko, Mikhail Malyuk, Andrey Holmanyuk, and Nikolai Liubimov,
\newblock ``{Label Studio}: Data labeling software,'' 2020-2022,
\newblock Open source software available from
  https://github.com/heartexlabs/label-studio.

\bibitem{basaldella-etal-2020-cometa}
Marco Basaldella, Fangyu Liu, Ehsan Shareghi, and Nigel Collier,
\newblock ``{COMETA}: A corpus for medical entity linking in the social
  media,''
\newblock in {\em Proceedings of the 2020 Conference on Empirical Methods in
  Natural Language Processing (EMNLP)}, Bonnie Webber, Trevor Cohn, Yulan He,
  and Yang Liu, Eds., Online, Nov. 2020, pp. 3122--3137, Association for
  Computational Linguistics.

\bibitem{9039685}
Jing Li, Aixin Sun, Jianglei Han, and Chenliang Li,
\newblock ``A survey on deep learning for named entity recognition,''
\newblock {\em IEEE Transactions on Knowledge and Data Engineering}, vol. 34,
  no. 1, pp. 50--70, 2022.

\bibitem{liu2021self}
Fangyu Liu, Ehsan Shareghi, Zaiqiao Meng, Marco Basaldella, and Nigel Collier,
\newblock ``Self-alignment pretraining for biomedical entity representations,''
\newblock in {\em Proceedings of the 2021 Conference of the North American
  Chapter of the Association for Computational Linguistics: Human Language
  Technologies}, June 2021, pp. 4228--4238.

\bibitem{pubmedbert}
Yu~Gu, Robert Tinn, Hao Cheng, Michael Lucas, Naoto Usuyama, Xiaodong Liu,
  Tristan Naumann, Jianfeng Gao, and Hoifung Poon,
\newblock ``Domain-specific language model pretraining for biomedical natural
  language processing,'' 2020.

\bibitem{jin2023medcpt}
Qiao Jin, Won Kim, Qingyu Chen, Donald~C Comeau, Lana Yeganova, W~John Wilbur,
  and Zhiyong Lu,
\newblock ``Medcpt: Contrastive pre-trained transformers with large-scale
  pubmed search logs for zero-shot biomedical information retrieval,''
\newblock {\em Bioinformatics}, vol. 39, no. 11, pp. btad651, 2023.

\bibitem{https://doi.org/10.48550/arxiv.2112.07869}
Robert Tinn et~al.,
\newblock ``Fine-tuning large neural language models for biomedical natural
  language processing,'' 2021.

\bibitem{snobert}
Mikhail Kulyabin et~al.,
\newblock ``A benchmark for clinical notes entity linking in the snomed ct
  clinical terminology,'' \url{https://github.com/MikhailKulyabin/SNOBERT}.

\bibitem{ayoola-etal-2022-refined}
Tom Ayoola, Shubhi Tyagi, Joseph Fisher, Christos Christodoulopoulos, and
  Andrea Pierleoni,
\newblock ``{R}e{F}in{ED}: An efficient zero-shot-capable approach to
  end-to-end entity linking,''
\newblock in {\em Proceedings of the 2022 Conference of the North American
  Chapter of the Association for Computational Linguistics: Human Language
  Technologies: Industry Track}, Anastassia Loukina, Rashmi Gangadharaiah, and
  Bonan Min, Eds., Hybrid: Seattle, Washington + Online, July 2022, pp.
  209--220, Association for Computational Linguistics.

\bibitem{kweon2023publicly}
Sunjun Kweon et~al.,
\newblock ``Publicly shareable clinical large language model built on synthetic
  clinical notes,'' 2023.

\end{thebibliography}

\end{document}